\documentclass[10pt,twocolumn,letterpaper]{article}

\usepackage{cvpr}      
\usepackage{adjustbox}
\usepackage[table]{xcolor}
\usepackage{tabu}
\usepackage{amsmath}
\DeclareMathOperator*{\argmax}{arg\,max}

\definecolor{cvprblue}{rgb}{0.21,0.49,0.74}
\definecolor{orange}{rgb}{1.0, 0.75, 0.0}
\definecolor{green}{rgb}{0.0, 0.5, 0.0}
\definecolor{amber}{rgb}{1.0, 0.30, 0.0}

\usepackage[pagebackref,breaklinks,colorlinks,allcolors=cvprblue]{hyperref}
\usepackage{multirow}
\usepackage{booktabs} 

\usepackage{arydshln}
\newcommand{\ApproachName}
{SEAL}

\def\logo{\makebox[0pt][l]{\hspace{0pt}\raisebox{-0.3ex}{\includegraphics[height=24pt]{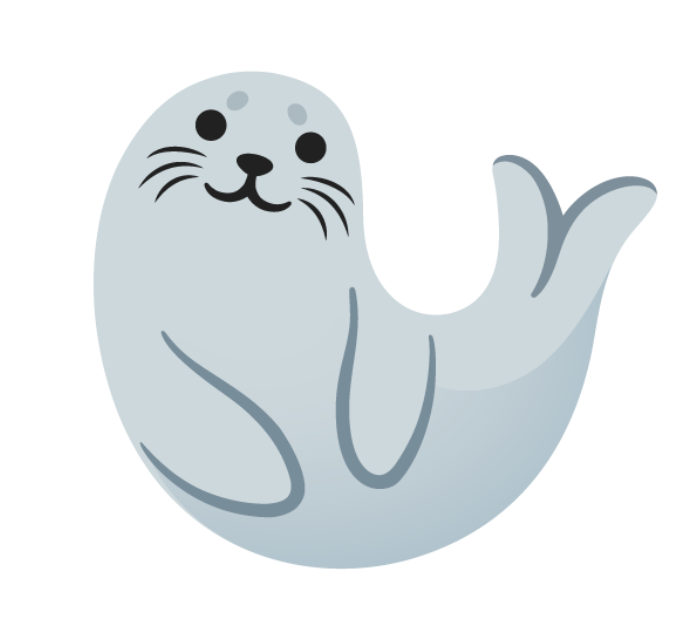}}}}

\title{\logo \ \ \ \ \ \ \ SEAL: \underline{SE}mantic \underline{A}ttention \underline{L}earning for Long Video Representation}

\author{Lan Wang$^{1,2}$\thanks{This work was done as Lan Wang’s internship project at Google.} \hspace{6mm} Yujia Chen$^2$ \hspace{6mm} Du Tran$^2$ \hspace{6mm}Vishnu Naresh Boddeti$^1$\hspace{6mm}~Wen-Sheng Chu$^2$ \\
$^1$ Michigan State University \hspace{29mm} $^2$ Google \\
{\tt\small \{wanglan3, vishnu\}@msu.edu \hspace{4mm} \{yujiachen, tranldu, wschu\}@google.com}
}

\begin{document}
\maketitle
\begin{abstract}

Long video understanding presents challenges due to the inherent high computational complexity and redundant temporal information. 
An effective representation for long videos must efficiently process such redundancy while preserving essential contents for downstream tasks. 
This paper introduces {\bf SE}mantic {\bf A}ttention {\bf L}earning (\ApproachName), a novel unified representation for long videos. 
To reduce computational complexity, long videos are decomposed into three distinct types of semantic entities: scenes, objects, and actions, allowing models to operate on a compact set of entities rather than a large number of frames or pixels. 
To further address redundancy, we propose an attention learning module that balances token relevance with diversity, formulated as a subset selection optimization problem. 
Our representation is versatile and applicable across various long video understanding tasks. 
Extensive experiments demonstrate that \ApproachName~significantly outperforms state-of-the-art methods in video question answering and temporal grounding tasks across diverse benchmarks, including LVBench, MovieChat-1K, and Ego4D.

\end{abstract}
    
\vspace{-2ex}
\section{Introduction}
\label{sec:intro}
\vspace{-0.5ex}

\begin{figure*}
    \centering
    \includegraphics[width=0.96\textwidth]{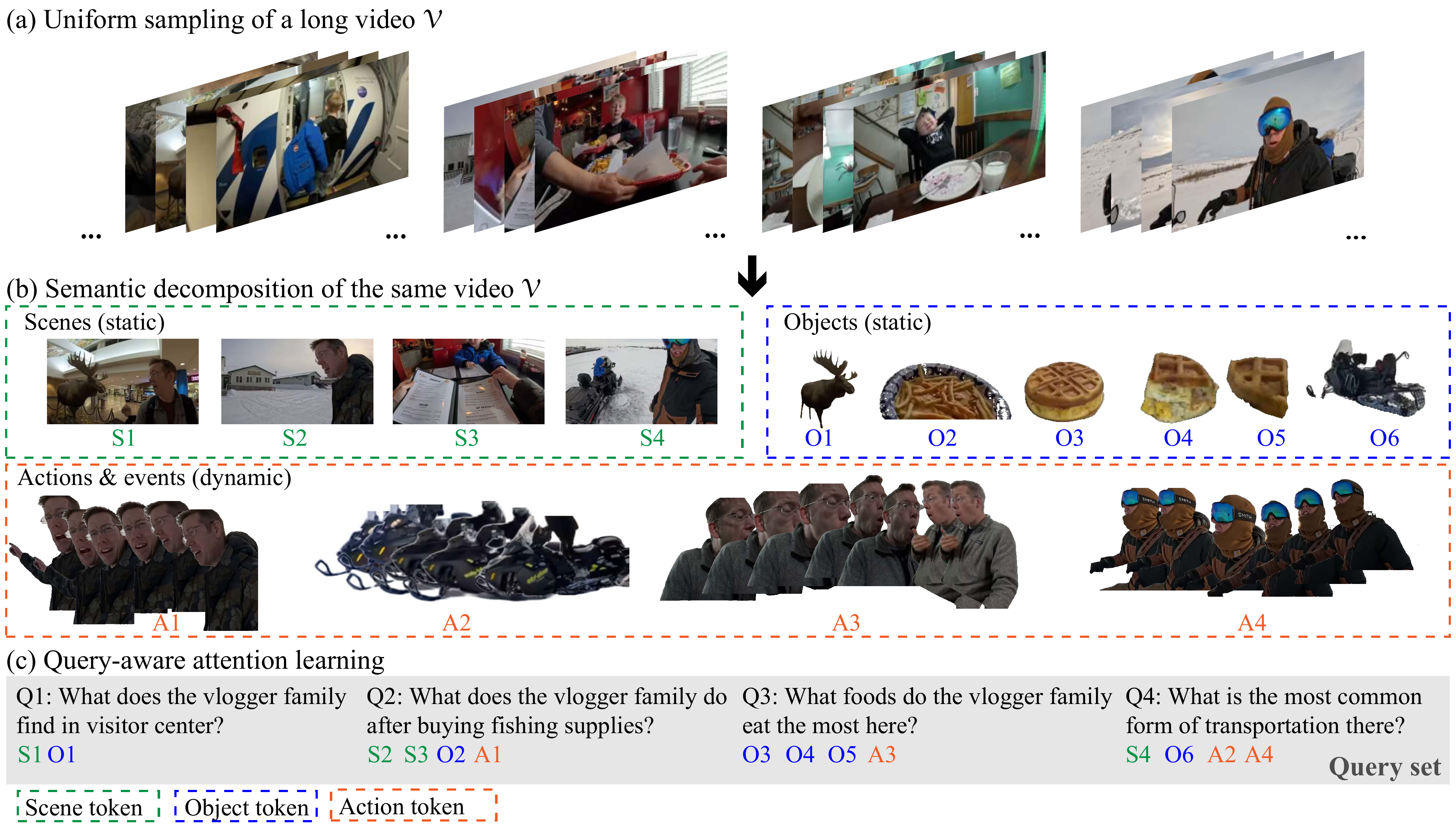}
    
    \vspace{-8pt}
    \caption{{\bf Long Video Representation with Semantic Attention Learning (\ApproachName):} 
    (a) Conventional uniform sampling results in redundant and cluttered visual information, making it difficult for both AI models and human brains to process efficiently. 
    (b) Decomposing long videos into semantic entities such as scenes, objects, and actions reduces temporal redundancy, thus making model training and inference more efficient.
    In this example, the long video $\mathcal{V}$ is decomposed into 4 {\color{green}{scene tokens (S1--S4)}}, 6 {\color{blue}{object tokens (O1--O6)}}, and 4 {\color{amber}{action/event tokens (A1--A4)}}.
    (c) Query-aware attention learning module improves downstream task performance by focusing on relevant information rather than processing everything.
    Queries (Q1--Q4) are shown with their most relevant tokens. (best viewed in color)}
    \vspace{-2ex}
    \label{fig:teaser}
\end{figure*}

State-of-the-art video understanding models excell at short video tasks such as video classification~\cite{kinetics,goyal2017something}, temporal grounding~\cite{wang2023protege} and action detection~\cite{thumos,caba2015activitynet}, which involve videos lasting from {\em a few seconds} to {\em minutes}.
However, their performance declines on hour-long videos~\cite{wang2024lvbench,weng2025longvlm}.
In contrast, a 10-year-old child can watch a full-length movie ({\em one to two hours}) and effortlessly answer questions at various levels of detail. 
This disparity between humans and machines emphasizes the foundamental challenges in long video understanding for AI models, including:
(1) {\bf Increased complexity}: Long videos require more computation and memory than current hardware can support for training or inference, 
(2) {\bf Temporal redundancy}: Slow-changing scenes and objects introduce significant redundancy, and 
(3) {\bf Cross-task generalization}: A robust representation must adapt to various tasks, from fine-grained fact retrieval to high-level reasoning. 
These challenges, which appear trivial for a child, remain  challenging for AI models. 

How does the human brain process long videos, particularly in addressing the above-mentioned challenges? 
First, rather than processing every pixel or frame, the brain selectively attends to new information to efficiently manage temporal redundancy \cite{Das1556,Luck1998-LUCOTR}.
Second, humans process videos in an online fashion, continuously updating their understanding and memories as they watch, rather than deferring reasoning until the end.
This continuous knowledge update, combined with selective attention, allows the brain to efficiently handle the complexity of long videos.
Finally, attention dynamically shifts based on context. 
Without specific guidance, a child may focus on naturally engaging or memorable moments.
However, when given specific questions in advance, attention becomes goal-oriented to seek relevant details while maintaining a broad understanding.
This suggests effective representations should balance between task-specific focus and holistic understanding of the video to enable cross-task generalization.

Inspired by how humans process long videos, we introduce {\bf SE}mantic {\bf A}ttention {\bf L}earning (\ApproachName), a novel unified representation designed to tackle the three key challenges in long video understanding. 
\ApproachName{}~consists of two main steps: {\em Semantic Decomposition} and {\em Attention Learning}.
In {\em Semantic Decomposition}, long videos are decomposed into three semantic entities such as scenes, objects, and actions, which are then treated as tokens. 
While the scene and object tokens represent static content assumed not to change rapidly, the action tokens are designed to capture the dynamic, fast-changing moments of the video. 
We note that these semantic tokens efficiently encode the essential information needed to answer ``where'', ``what'', or ``how'' questions about the videos. 
This decomposition significantly reduces complexity by allowing AI models to operate on a compact set of tokens instead of raw pixels or frames. 
Figure~\ref{fig:teaser}(a) illustrates a conventional uniformly sampled video $\mathcal{V}$, where redundant frames create cluttered visual information that hinders effective analysis for both models and humans. 
Figure~\ref{fig:teaser}(b) shows our semantic decomposition, breaking down the video into distinct scenes, static objects, and dynamic actions.
In Attention Learning, we formulate a subset selection problem that maximizes the query relevance while ensuring token diversity.
This step not only mitigates redundancy but also enhances cross-task generalization by prioritizing the most informative tokens.
Figure~\ref{fig:teaser}(c) illustrates our attention learning module with four different queries and their selected tokens that capture relevance and diverse video content. 
Finally, \ApproachName{} is designed to work for both {\em global} and {\em streaming} modes, enabling it to process arbitrarily long videos. 
Extensive ablations and experiments confirm \ApproachName's superior performance over existing methods. 
Our key contributions include:
\begin{itemize}
    \item We introduce \ApproachName, a novel unified representation for long videos by decomposing them into three semantic tokens, namely scenes, objects, and actions.
    
    \item Our attention learning module reduces temporal redundancy while supporting strong cross-task generalization.
    We show \ApproachName~works in both global and streaming modes, making it adaptable to arbitrarily long videos.
    
    \item \ApproachName outperforms state-of-the-art methods on various long video understanding tasks and benchmarks including: video QA (MovieChat-1K~\cite{song2023moviechat}, LVBench~\cite{wang2024lvbench}), and egocentric video grounding (Ego4D~\cite{Ego4D}).
\end{itemize}
\section{Related Work}
\label{sec:relatedwork}
\vspace{-0.5ex}

Recent approaches have unified various video understanding tasks by framing them as video QA tasks~\cite{wang2024lvbench}, leveraging the capabilities of LLMs. However, fine-tuning with task-specific vision heads continues to offer advantages in memory efficiency and task-specific performance~\cite{pei2024egovideo}, particularly for temporal grounding. In this section, we review advancements in Video Question Answering (QA) and temporal grounding for long video understanding.

\noindent \textbf{Video QA for long videos}.
The main challenge for long video QA is the memory constraint.
He~\etal~\cite{he2024ma} introduced a sequential framework that uses a memory bank to enhance long-term comprehension. Song~\etal~\cite{song2023moviechat,song2024moviechat+} integrated video foundation models with LLMs through a memory mechanism inspired by the Atkinson-Shiffrin model, reducing computational complexity while preserving long-term memory. Another line of work improves efficiency by decomposing video content. Rui~\etal~\cite{qian2024streaming} employed Memory-Propagated Streaming Encoding to segment videos into short clips, with Adaptive Memory Selection enhancing response accuracy by identifying question-relevant memories. Min~\etal~\cite{min2024morevqa} introduced a multi-stage, training-free framework, emphasizing task decomposition into parsing, grounding, and reasoning stages. More recently, Weng~\etal~\cite{weng2025longvlm} proposed a hierarchical framework that encodes local features and integrates global semantics for detailed comprehension of extended video content. To further reduce model's hallucination in QA, Sun~\etal~\cite{sun2024hallucination} proposed a question-guided pipeline by focusing on relevant frames and controlled answer generation.

\noindent \textbf{Temporal localization for long videos}. Recent research in temporal localization for long videos has explored two primary directions: \textbf{LLM-based} and \textbf{non-LLM-based} approaches. With the advances in LLMs, researchers have expanded its use beyond traditional VQA, leveraging the capabilities for temporal grounding tasks, with a primary focus on enhancing localization accuracy.
Ren~\etal~\cite{ren2024timechat} proposed a timestamp-aware model that aligns visual content with temporal cues, enabling adaptive processing of sequential events for tasks like localization.
Similarly, Fan~\etal~\cite{fan2025videoagent} introduced a memory-enhanced framework that captures contextual information across video segments, improving temporal and spatial reasoning.
Korbar~\etal~\cite{korbar2025text} developed a text-guided resampling mechanism that dynamically selects video segments, focusing on relevant scenes to enhance temporal and spatial comprehension.

In contrast, non-LLM-based approaches typically rely on training a regression layer or decoder for temporal localization, with efforts concentrated on refining visual features and multimodal fusion to improve alignment. For instance, Hou~\etal~\cite{hou2022cone} introduced a hierarchical framework that combines coarse scanning with fine-grained alignment to optimize both precision and efficiency in localizing target moments. Pan~\etal~\cite{pan2023scanning} applied a coarse-to-fine pipeline for single-pass temporal grounding that improves both efficiency and alignment. 
Additionally, Mu~\etal~\cite{mu2024snag} proposed a cost-effective late fusion approach paired with a video-centric sampling scheme to improve scalability.

Unlike prior works that focus on specific tasks, \ApproachName~proposes a unified and generic framework for long video understanding, capable of adapting to various prediction heads and tasks. Additionally, \ApproachName~supports both global and streaming modes, making it adaptable to arbitrarily long videos.
\section{Semantic Decomposition and Attention Learning}
\label{sec:method}
\begin{figure*}
    \centering
    \includegraphics[width = 0.9\textwidth]{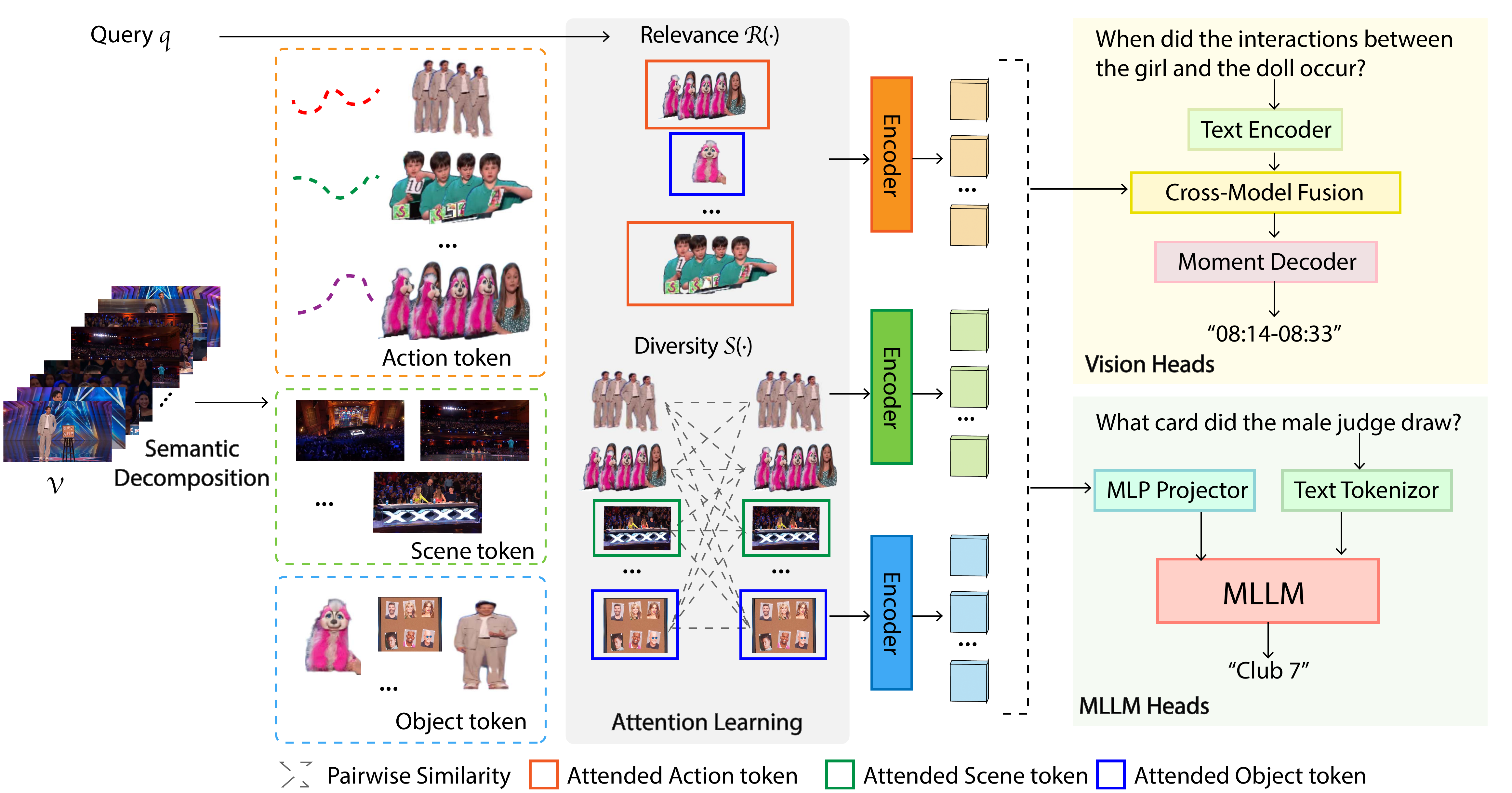}
    \vspace{-8pt}
    \caption{{\bf \ApproachName~Overview}. During {\em semantic decomposition}, a long video $\mathcal{V}$ is decomposed into semantic tokens representing scenes, objects, and actions. Then, during {\em attention learning}, these tokens and the query q, are optimized for query relevance $R(\cdot)$ and token diversity $S(\cdot)$. 
    The resulting attended token subset is then passed to a vision or an MLLM head for predictions.} 
    \vspace{-1ex}
    \label{fig:lvu_overview}
\end{figure*}

Let $\mathcal{V}=\{v_i\}_{i=1}^{T_V}$ be an arbitrarily long untrimmed video, where $v_1, \hdots, v_{T_V}$ denote the sequence of $T_V$ frames forming the video. 
Let $q$ be a query from long video understanding tasks, comprising a sequence of $l_q$ tokens. 
The query $q$ may take different forms depending on the task, such as natural language text for video question and answering (\eg, MovieChatQA~\cite{song2023moviechat}, LVBench~\cite{wang2024lvbench}) or visual/text  template or action label for episodic memory tasks in egocentric video understanding (\eg, Ego4D~\cite{Ego4D}). 
Our method establishes a unified video representation to generalize across these diverse long video understanding tasks. An overview of our approach is presented in Figure~\ref{fig:lvu_overview}.

\subsection{Semantic Decomposition of Long Videos}

The main challenge in long video representation lies in capturing diverse content within limited memory. 
Conventional methods resort to frame sampling~\cite{Zhang2023VideoLLaMAAI,xu2024pllava} or maintaining a memory bank that merges similar frames~\cite{song2023moviechat,song2024moviechat+}. 
However, these approaches can vary greatly across tasks. We propose a novel decomposition approach that structures long videos into three distinct token types representing different type of semantic entities:
(1) Scene tokens $\mathbf{T}_{\text{scene}}$ capture background context, providing essential cues about the environment.
(2) Object tokens $\mathbf{T}_{\text{object}}$, highlight key static elements relevant to specific tasks. 
(3) Action tokens $\mathbf{T}_{\text{action}}$ represent moving elements, focusing on temporal information such as motions, activities, or events.
With this structured tokenization, we create a unified, compact, task-agnostic representation that minimizes the need for redundant and dense video storage while preserving a comprehensive understanding of long video content.

\noindent {\bf Scene Tokens.} 
Any frame in a video can serve as a scene token because it captures the environment where it was recorded, \eg, indoor gym, outdoor mountain, etc.
Although shot boundary detection could split a long video into shots with one scene token per shot, this approach has two drawbacks:
(1) Shot detection algorithms are often imperfect, with no opportunity to correct once shot boundary is determined.
(2) Shot detection often ignores the specific query and downstream tasks, and thus could cause sub-optimal performance.
To overcome these drawbacks, we propose a two-step approach.
First, we over-sample scene tokens, and then later perform an attention learning step to maximize query relevance and token diversity. 
Specifically, we uniformly sample \(N_{\text{scene}}\) frames to capture a diverse background. 
These pre-sampled scene tokens, denoted as $\mathbf{T}_{\text{scene}}=[t^{scene}_i]_{i=1}^{N_{\text{scene}}}$, will undergo another round of attention learning, detailed in Section~\ref{sec:adaptive_sampling}.

\noindent \textbf{Action Tokens.} The purpose of the action token is to capture temporal information of moving objects such as low-level motions, activities, and events. We begin by using a class-agnostic object tracker, \eg, SAM-2~\cite{SAM2}, to extract multiple initial dynamic tracklets. 
Tracklets shorter than \(L_{\text{min}}\) are discarded, while those longer than \(L_{\text{max}}\) are split into multiple tracklets with length of \(L_{\text{max}}\).
For each tracklet, we take the spatial union of bounding boxes across frames, allowing the dynamic token to capture not only motion information but also the spatial movement of people and objects.
This process yields \(N_{\text{tracklet}}\) tracklets, denoted as $\mathbf{T}_{\text{action}} = \{\tau_{\text{dynamic}}^i\}_{i=1}^{N_{\text{tracklet}}}$ .

\noindent \textbf{Object Tokens.} For the object tokens, we utilize a class-agnostic grouping method such as SAM~\cite{SAM} to generate masks for all objects in each frame. This class-agnostic segmentation approach enables comprehensive object information capture. Specifically, we apply this process on \(N_{\text{key}}\) key frames. On each of the \(N_{\text{key}}\) frames, we apply SAM to obtain in total \(N_{\text{object}}\) object masks, which we denote as \(\mathbf{T}_{\text{object}} = \{\mathbf{M}_{\text{object}}^i\}_{i=1}^{N_{\text{object}}}\). 
Key frame selection varies by task and is detailed in the following sections.

\subsection{Attention Learning}
\label{sec:adaptive_sampling}

Although the long video is decomposed into different token representations, the resulting tokens are still redundant. To address this, we propose a sampling approach that balances between query-relevance and token-diversity, denoted as Attention Learning. Specifically, we formulate our sampling as an optimization problem with two main objectives: query-relevance and token-diversity.
\begin{equation}
\begin{split}
    T_s^*& = \argmax_{T_s \subset T_G} F_s(T_s | T_G, q)  \\
    &= \argmax_{T_s \subset T_G} \alpha \!\!\sum_{t_s \in T_s} \!\!R(t_s, q) \!+\! (1\!-\!\alpha) \!\!\!\!\!\sum_{t_i, t_j \in T_s, i \neq j} \!\frac{1}{S(t_i, t_j)}. \nonumber
    \label{eq:sampling}
\end{split}
\end{equation}

We aim to solve a subset selection problem: find a fixed-size subset $T_s \!\subset\! T_G$ that maximizes the objective function $F_s(T_s|T_G,q)$. Here, $T_G$ represents the set of all tokens used, which can be any set of tokens such as scene $\mathbf{T}_{\text{scene}}$, object $\mathbf{T}_{\text{object}}$, action $\mathbf{T}_{\text{action}}$, or a combination of all ($\mathbf{T}_{\text{scene}} \cup \mathbf{T}_{\text{object}} \cup \mathbf{T}_{\text{action}}$). 
$q$ denotes the query being asked by the downstream understanding task. And, $T_s^*$ denotes the optimal subset that maximizes $F_s$. 
$F_s$ is decomposed into two terms: $R(\cdot)$ measures the relevance between a visual token and the query which we compute by encoding them with the BLIP-2 model~\cite{li2023blip} and calculating their cosine similarity; and $S(\cdot)$ calculates the cosine similarity of the paired tokens. We note that the first term maximizes the relevance between selected tokens and the query while the second term enforces token diversity (via minimizing the token pairwise similarity). Finally, $\alpha$ is a hyper-parameter to balance between query-relevance and token-diversity. 

\vspace{-0.5ex}
\subsection{Streaming and Global Mode}

Our proposed method offers two ways of representing long videos: \textit{streaming} and \textit{global}. In the global mode, the model fully ``watches'' (or processes) the entire video and then provides a single representation. In contrast, in the streaming mode, the model processes the video buffer by buffer and provides an updated representation at any given time step. The partly-observed video representation can change over time as the video progresses. This setting simulates the situation when you watch a movie together with a child and have interactions with her or him as the movie is still going on.

In the global model, all tokens undergo a single optimization through Attention Leaning, resulting in \( T_{\text{sub}} \), which includes \( k \) tokens. This $T_{\text{sub}}$ is regarded as the representation for the entire video. We note that the global mode may not scale well with arbitrary long videos because all tokens cannot be fit into a limited memory for sampling. One can opt to use more aggressive uniform temporal pre-sampling to reduce the numbers of tokens before Attention Learning. This workaround can bypass the memory limitation, but may also lead to sub-optimal solutions due to missing important tokens due to uniform sampling.

Alternatively, we propose an online streaming approach for representing partly-observed videos. Specifically, we use a fixed-size sliding window with a size set to the maximum number of tokens $l$ allowed by memory capacity, denoted as $T$. At each step $t$, we apply Attention Learning to the union set of the tokens in the current window $T_t$ and the previous selected subset of tokens $T_{\text{sub}}^{t-1}$ and obtain the representation of the video at time $T_{\text{sub}}^t$. At the beginning, the selected subset is set to empty ($T_{\text{sub}}^{0}=\emptyset$).  
\begin{equation}
T_{\text{sub}}^t = \text{Attention\_Learning}(T_{t} \cup T_{\text{sub}}^{t-1}) \ \forall t>0.
\end{equation}

This streaming mode allows us to use $T_{\text{sub}}^t$ as the partly-observed representation of the video and can be fed into any prediction head for video understanding tasks. As an immediate benefit of the streaming mode, our proposed representation now can handle arbitrary long videos.

\subsection{Prediction Heads}

Our unified representation is adaptable to most long video understanding tasks using different prediction heads. In this paper, we demonstrate two specific use cases of our representation: one is used with the traditional vision head for video temporal grounding and the other one is with the Multimodal LLMs (MLLM) head for video QA.

\noindent {\bf Temporal Grounding with Vision Heads}. Given a query \( q \), the task is to locate the start and end times, \( t_{\text{start}} \) and \( t_{\text{end}} \), where the answers could be deduced. We first encode the sampled tokens \( T_{\text{sub}} \) and the query using encoders \( \mathbb{E}_V \) and \( \mathbb{E}_q \) to obtain embeddings for each video token \(
 z_v^i = \mathbb{E}_V(T_{\text{sub}^i})
\) and query
 \(
 z_q = \mathbb{E}_q(q)
\).
The cross-modal fusion is then performed to obtain the fused representation:

\begin{equation}
z_{\text{joint}}^i = \text{CrossModalFusion}(z_v^i, z_q)
\end{equation}
Finally, a moment decoder is applied to predict the start and end time for the query \( q \). The moment decoder includes a classification head and a regression head. The classification head is used to predict the score \(p_{score}^i\) of each token, while the regression head predicts the normalized distances \((d_{\text{start}}^i, d
_{\text{end}}^i)\) from each token to the moment boundaries:

\begin{equation} p_{score}^i,
(d_{\text{start}}^i, d
_{\text{end}}^i) = \text{MomentDecoder}(z_{\text{joint}}^i).
\end{equation}
The regression and classification heads are optimized using an IoU distance and a focal loss as used in~\cite{MuM024}. The final moment is calculated as $t_{start}^i,t_{end}^i = (t_i-d_{\text{start}}^i) \times L_V,(t_i+d_{\text{end}}^i) \times L_V$, where \(L_V\) is the length of the input video.
The proposed unified representation enables efficient handling of long video sequences and allows for localization related tasks.

\noindent {\bf Video QA using MLLM Heads}. The proposed representation can also be connected to an MLLM head, making it applicable to various video QA-related tasks, such as reasoning, understanding, and summarization. Moreover, some grounding tasks can also be addressed in a QA format. Specifically, \( z_v \) is projected through an MLP to a visual token \(T_v^{\text{MLLM}} \) that the MLLM can interpret, which is then input into the MLLM along with text tokens \(T_{text}\). Based on different benchmarks, the MLLM performs multiple-choice or open-ended answering.

\vspace{-1ex}
\section{Experiments}
\label{sec:experiment}
\vspace{-0.5ex}

\begin{table*}[ht!]\small
\centering
\begin{tabular}{lcccccccc}
\toprule
\textbf{Model} & \textbf{LLM Size} & \textbf{Overall (\%)} & \textbf{KIR (\%)} & \textbf{EU (\%)} & \textbf{Sum (\%)} & \textbf{ER (\%)} & \textbf{Rea (\%)} & \textbf{TG (\%)}  \\
\midrule
\rowcolor{gray!10}
Qwen2-VL~\cite{Qwen2VL} & 72B & \underline{41.3} & 38.3 & \underline{41.1} & \textbf{46.6} & \underline{38.0} & \textbf{46.5} & \textbf{41.4} \\
\midrule
InternVL2~\cite{internvl} & 34B & 39.6 & \underline{43.4} & 39.7 & \underline{41.4} & 37.4 & 42.5 & 31.4 \\
LLaVA-NeXT~\cite{zhang2024llavanextvideo} & 34B& 32.2 & 34.1& 31.2 & 27.6 & 30.1 & 35.0 & 31.4  \\
Oryx~\cite{liu2024oryx} & 34B & 30.4 & 32.1 & 29.2 & 27.6 & 30.1 & 34.0 & 29.1\\
PLLaVA~\cite{xu2024pllava} & 34B & 26.1  & 26.2 & 24.9 & 25.9 & 25.0 & 30.0 & 21.4\\
\midrule
\rowcolor{orange!20}
\textbf{SEAL (Ours)} & 34B & \textbf{45.9} & \textbf{51.5} & \textbf{41.3} & 39.7 & \textbf{47.9} & \underline{43.3} & \underline{32.3} \\

\bottomrule
\end{tabular}
\vspace{-8pt}
\caption{{\bf Comparison with state-of-the-art on LVBench}. Our method achieves the highest overall score, with notable gains in Key Information Retrieval (KIR) and Entity Recognition (ER), demonstrating the effectiveness of our representation in locating key information and entities by eliminating redundancy. The best methods are highlighted in \textbf{bold}, and the second-best are \underline{underlined}.}
\vspace{-2ex}
\label{tab:lvbench-quant}
\end{table*}

\subsection{Implementation Details}

\noindent \textbf{Datasets and metrics}. We evaluate \ApproachName~on three datasets, each selected for its relevance to long video understanding on different capabilities. {\em LVBench}~\cite{wang2024lvbench} contains 1,549 QA pairs across six tasks, with videos averaging 4,101 seconds (approximate 1 hour 8 minutes). Each question presents a single-choice format with four options. Accuracy serves as the evaluation metric for individual tasks as well as overall performance across all tasks. We primarily focus on this dataset due to its emphasis on hour-long videos. {\em Moviechat-1K}~\cite{song2023moviechat} includes 1,000 video clips with dense captions spanning 15 categories, averaging 564 seconds (about 10 minutes). The benchmark employs LLMs, specifically GPT-3.5~\cite{openai2023gpt35}, to evaluate the quality of generated answers. A rating ranging from 0 to 5 is used to compute the overall score, while a binary preference from the LLM is used to calculate the accuracy. {\em Ego4D-NLQ}~\cite{Ego4D} is a part of the Ego4D Episodic Memory challenge for Natural Language Queries (NLQ) task. This dataset requires localizing a temporal window where the answers can be deduced from untrimmed egocentric videos. It contains 1,259 videos, averaging 10 minutes each. Our experiments apply memory constraints to simulate long video scenarios. Metrics include Top-1 and Top-5 recall at various thresholds.

\noindent \textbf{Experiment setup}. We fine-tune the projection layers and Q-former~\citep{li2023blip} to adapt to different types of tokens for 20 epochs using the training split of MovieChat-1K.
For Ego4D-NLQ, we use the training split to finetune the vision heads for 7 epochs. The AdamW optimizer~\cite{loshchilov2019adamw} is employed with default beta values of ($0.9$, $0.999$) and a weight decay of $0.05$.
For token extraction, we use SAM2~\cite{SAM2} to obtain object tokens, YOLOv10-X~\cite{Wang2024YOLOv10RE} with BoT-SORT~\cite{Aharon2022BoTSORTRA} for action tokens. Scene, action, and object tokens are extracted at 8, 10, 1 FPS for MovieChat-1K, Ego4D-NLQ and LVbench. For the LVBench, we follow the settings of InternVL2 and utilize Yi-34B~\cite{Young2024YiOF}. For MovieChat-1K, we adopt the same settings as \cite{song2023moviechat} and use Vicuna-7B as~\cite{song2023moviechat}. For Ego4D-NLQ, we adhere to the settings specified in SnAG~\cite{MuM024} and use EgoVLP~\cite{kevin2022egovlp} as vision encoder. We use $0.9$ as the default value for the only hyper-parameter $\alpha$.

\begin{table}[t] 
\footnotesize\centering
\begin{tabu} to 0.45\textwidth {@{}l*{7}{c}@{}}
\toprule
& & \multicolumn{3}{c}{\textbf{R@1}} &  \multicolumn{3}{c}{\textbf{R@5}} \\
\cmidrule(lr){3-5} \cmidrule(lr){6-8}
\textbf{Model} & \textbf{\#Token} & 0.3 & 0.5 & Avg & 0.3 & 0.5 & Avg \\
\midrule
SnAG & all & 15.72 & 10.78 & 13.25  & 38.39  & 27.44 & 32.92\\
\midrule
SnAG & 450  & 13.44 & 9.23 & 11.34 & 34.02 & 23.04 & 28.53 \\
\textbf{SEAL} & 450 & \textbf{13.78} & \textbf{9.26} & \textbf{11.52} & \textbf{34.79} & \textbf{23.10} &  \textbf{28.95}\\
\midrule
SnAG & 200  & 10.03 & 6.35 & 8.19 & 26.56 & 16.90 & 21.73 \\
\textbf{SEAL} & 200 & \textbf{10.83} & \textbf{7.06} & \textbf{8.95} & \textbf{27.39} & \textbf{17.41} & \textbf{22.40} \\
\bottomrule
\end{tabu}
\vspace{-8pt}
\caption{{\bf Comparisons with SoTA methods on Ego4D-NLQ}. 
Quantitative results for temporal grounding on Ego4D episodic memory Natural Language Queries (NLQ) task show our method, \ApproachName, consistently outperforms SnAG~\cite{MuM024} in all metrics under memory constraints with varying number of tokens. 
}
\label{tab:ego4d-quant}
\vspace{-1.5ex}
\end{table}

\subsection{Comparison with State-of-the-arts}
We conduct quantitative analysis on LVBench dataset which contains videos on lengths averaged more than 1 hour. Additionally, we demonstrate the generalizability of the proposed unified representation using MovieChat-1K, which has open-ended QA questions on a variety of scenarios, and Ego4D-NLQ, where we use traditional vision decoders (LLM-independent) for the temporal grounding task.

\noindent {\bf LVBench}. Table~\ref{tab:lvbench-quant} shows the accuracy of our method on different categories of LVBench dataset~\cite{wang2024lvbench}, demonstrating that \ApproachName~achieves the highest overall score, notably outperforming even larger models like Qwen2-VL-72B~\cite{Qwen2VL} by 4.6\%. Our method excels in Key Information Retrieval (KIR) and Entity Recognition (ER), outperforming the strongest alternatives by 8.1\% and 5.1\% respectively.
These results highlight that our action and object tokens effectively locate key information and entities by removing redundancies. While larger LLMs often exhibit better performance, this analysis suggests that it is not the only determinant, and a unified representation like ours can achieve state-of-the-art results with fewer parameters.

\begin{figure*}
    \centering
    \includegraphics[width = 0.85\textwidth]{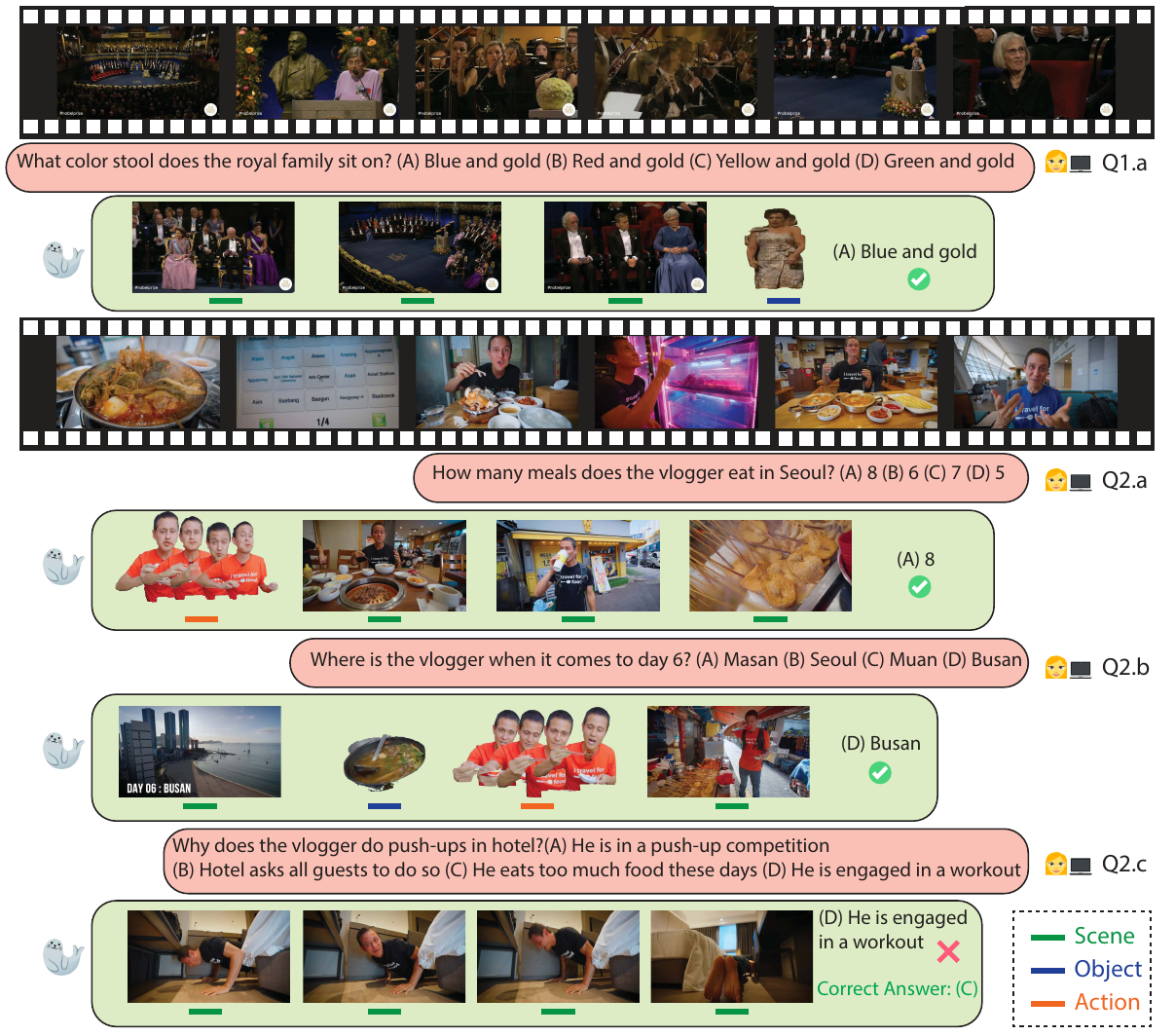}
    \vspace{-8pt}
    \caption{{\bf Qualitative results on LVBench}. Two long videos visualized with questions, multiple choice options, and \ApproachName~predicted answers. \ApproachName~ attends to relevant entities such as ``royal family'' and ``stool'' (Q1.a), different ``meals'' and ``drinks'' (Q2.a), ``scene'' and ``location'' (Q2.b) and correctly answers these questions. Although attending to relevant ``push-up'' activity (Q2.c), \ApproachName~fails to predict the right answer due to the challenging in the causal reasoning question.}
    \label{fig:visualization}
    \vspace{-1.5ex}
\end{figure*}

Figure~\ref{fig:visualization} presents a qualitative analysis of \ApproachName~on LVBench, demonstrating that our approach is able to pay attention to relevant semantic tokens and make correct answers to different types of questions. \ApproachName~accurately locates relevant tokens, \eg, identifying the royal family’s stool color (Q1.a), counting meals eaten (Q2.a), or determining scene or location (Q2.b). We also show a failure case (Q2.c), where the nuanced ``why'' question requires complicated causal relations of different scenes.

\begin{table}[t]
    \small
    \centering
    \begin{adjustbox}{max width=0.5\textwidth}
    \begin{tabular}{llcc}
    \toprule
    & \textbf{Methods} & \textbf{Accuracy(\%)} & \textbf{Score}\\
    \midrule
    \multirow{4}{*}{\textbf{Zero-shot}} 
    & VideoChat~\cite{Li2023VideoChatCV} & 61.0 & 3.34  \\
    & VideoLLaMA~\cite{Zhang2023VideoLLaMAAI} & 51.4 & 3.10 \\
    & Video-ChatGPT~\cite{Maaz2023VideoChatGPTTD} & 44.2 & 2.71  \\
    & MovieChat~\cite{song2023moviechat} & 67.8 & 3.81  \\
    \midrule
    & TimeChat-Hal~\cite{sun2024hallucination} & 73.8 & 3.58  \\
    {\textbf{Supervised}} & HERMES~\cite{faure2024hermestemporalcoherentlongformunderstanding} & 84.9 & \textbf{4.40} \\
    & \textbf{SEAL (Ours)} & \textbf{86.8} & 4.35 \\
    \bottomrule
    \end{tabular}
    \end{adjustbox}
    \vspace{-8pt}
    \caption{{\bf Comparison with SoTA methods on MovieChat-1K}. \ApproachName~outperforms the second best method, HERMES, by 1.9\% on accuracy while being comparable on the score metric.}
    \vspace{-1ex}
    \label{tab:moviechat-quant}
    \vspace{-1.5ex}
\end{table}

\noindent {\bf Moviechat-1K}. Table~\ref{tab:moviechat-quant} compares our method with state-of-the-art methods on MovieChat-1K.
We focus on evaluating performance in the global mode, which assesses the model's ability to comprehend information from the entire video, rather than the breakpoint mode, which primarily evaluates its ability to answer questions related to specific timestamps.
Our method surpasses all the alternatives, achieving the highest accuracy and a strong score.
Notably, unlike LVBench, which employs multi-choice questions, this dataset evaluates generated answers using an off-the-shelf LLM. This highlights the ability of our learned representations to effectively transfer to downstream generative tasks, enabling generating accurate and detailed responses.

\noindent {\bf Ego4D-NLQ}. We also evaluate our approach on the Ego4D-NLQ task, which differs from previous datasets as it uses a decoder to locate events for temporal grounding, rather than relying on an LLM to generate text-based answers. 
Table~\ref{tab:ego4d-quant} shows the results under memory-constrained conditions, where we limit token numbers to simulate real-world memory limitations in long video understanding.
Our method consistently achieves the highest recalls across various thresholds. With further memory constraints (\eg, reducing tokens to 200), \ApproachName~outperforms the current SoTA method, SnAG~\cite{MuM024}, by a even larger margin.
This demonstrates the robustness of our unified representation on various downstream applications for long video understanding.

\subsection{Ablation Study}

\noindent {\bf Semantic decomposition}. We analyze the impact of the three proposed token types on LVBench (Table~\ref{tab:lvbench_tokentype}). To prevent LLMs from leveraging prior knowledge to ``guess'' answers using only the text query, we present the baseline that uses random token inputs, isolating the LLM's performance. Results indicate that using each token type individually significantly improves performance, highlighting the value of each token's contribution. Notably, scene tokens provide the most substantial improvement, aligning with prior research~\cite{tan2024koala,chen2022frame,weng2025longvlm} that emphasizes frame-based scene tokens. Furthermore, different token combinations yield additional gains: adding action tokens enhances Key Information Retrieval (KIR), Reasoning (Rea) and Temporal Grounding (TG) tasks by capturing temporal dynamics in the queries, while object tokens boost Entity Recognition (ER) performance by retaining detailed object-specific information.
We observe high variance in the Summarization (Sum) accuracy due to the small number of questions in this category and the randomness of LLM.
Similar improvements have been observed on the MovieChat-1K in Table~\ref{tab:moviechat_tokentype}, where the accuracy improved by 10.45\% with action and object tokens.
Ultimately, using all three token types together achieves the highest performance, demonstrating the complementary strengths of scene, action, and object tokens as a unified representation in long video understanding.

\begin{table}[tbh]
\footnotesize\centering
\begin{adjustbox}{max width=0.475\textwidth}
\begin{tabular}{lccccccc}
\toprule
\textbf{Model} & \textbf{Overall} & \textbf{KIR} & \textbf{EU} & \textbf{Sum} & \textbf{ER} & \textbf{Rea} & \textbf{TG} \\
\midrule
\rowcolor{gray!10}
Random tokens & 24.8 & 24.7 & 25.2 & 34.5 & 23.5 & 26.4 & 20.9 \\
\midrule
Action only & 34.0 & 33.3 & 31.7 & \underline{36.2} & 34.3 & 35.8 & 33.6 \\
Object only & 33.9 & 34.4 & 32.6 & 31.0 & 33.4 & 40.8 & 29.5 \\
Scene only & 42.8 & 50.5 & \underline{40.6} & 25.9 & 44.3 & 41.3 & 27.2 \\
\midrule
Scene+Object & 43.4 & 47.4 & 40.0 & 24.1 & \underline{47.6} & 42.3 & \underline{35.5} \\
Scene+Action & \underline{44.4}  & \textbf{53.3} & 40.0 & 24.1 & 46.2 & \textbf{43.3} & \textbf{38.2}\\
\midrule
\rowcolor{orange!20}
\textbf{SEAL (Ours)} & \textbf{45.9}  & \underline{51.5} & \textbf{41.3} & \textbf{39.7} & \textbf{47.9} & \textbf{43.3} & 32.3 \\
\bottomrule
\end{tabular}
\end{adjustbox}
\vspace{-8pt}
\caption{{\bf The effects of different types of tokens on LVbench}. Using any type of tokens outperforms the random baseline,
while applying all three types of tokens brings the best performance.}
\label{tab:lvbench_tokentype}
\vspace{-1ex}
\end{table}

\begin{table}[t]
\footnotesize\centering
\begin{tabu} to 0.47\textwidth {l*{4}{X[c]}}
\toprule
{\bf Method} & Scene & Scene+Object & Scene+Action & {\bf SEAL} \\
\midrule
Accuracy & 76.33 & 79.49 & 81.46 & {\bf 86.78} \\
Score & 4.15 & 4.26 & 4.28 & {\bf 4.35} \\
\bottomrule
\end{tabu}
\vspace{-8pt}
\caption{{\bf The effects of different types of tokens on MovieChat-1K}. Results are on the test set with our global inference mode. Using all three types of tokens provides the best accuracy.}
\label{tab:moviechat_tokentype}
\vspace{-1ex}
\end{table}

We observe that the latency bottleneck mostly comes from the tracker when processing multiple objects. Table~\ref{tab:lvbench-ablations} presents an analysis of different trackers for extracting action tokens. While YOLO+BoT-SORT operates in a much higher frame rate (14 FPS), it lacks open-set detection capabilities. On the other hand, SAM2 can detect all objects within a scene but operates significantly slower (8FPS). Our findings indicate that the proposed action tokens does not require capturing all the objects in the scene.
In practice, YOLO-X pretrained on COCO, with BoT-SORT, effectively captures essential information for action token extraction, maintaining a balance between performance and efficiency.

\noindent {\bf Attention learning}. Table~\ref{tab:lvbench-ablations} highlights the effectiveness of query relevance and token diversity terms in Section~\ref{sec:adaptive_sampling}. 
When $\alpha\!=\!0$, the optimization focuses on the diversity term to capture extensive contextual information for long-video tasks such as Summarization (Sum) and Reasoning (Rea), achieving reasonable results comparable to other state-of-the-art approaches in Table~\ref{tab:lvbench-quant}.
With $\alpha\!=\!1$, the optimization prioritizes the relevance term to reduce redundancy by sampling tokens that closely align with the query, so performance can be improved for tasks like Key Information Retrieval (KIR).
However, this approach loses global context and results in weaker summarization performance. 
Balancing both terms yields the best overall results.

\noindent {\bf Streaming mode}. We compare the global mode and streaming mode of our method on LVBench. As discussed in Section \S\ref{sec:method}, the streaming mode better simulates real-world scenarios by sequentially processing partial observations to aggregate information over arbitrarily long videos. Table~\ref{tab:lvbench-ablations} demonstrates that streaming mode performs worse on globally-dependent tasks such as Event Understanding (EU) and Key Information Retrieval (KIR). However, it excels in temporally-intensive tasks like Temporal Grounding (TG) and Reasoning (Rea). Notably, streaming mode still surpasses the most competitive baseline, even when the baseline uses a significantly larger LLM.

\begin{table}[t]
\footnotesize\centering
\begin{tabu} to 0.47\textwidth {@{}lX[2cm c]*{6}{X[c]}@{}}
\toprule
\textbf{Model} & \textbf{Overall} & \textbf{KIR} & \textbf{EU} & \textbf{Sum} & \textbf{ER} & \textbf{Rea} & \textbf{TG} \\
\midrule
{\bf SEAL (Ours)} & 45.9   & 51.5 & 41.3 & 39.7 & 47.9 & 43.3 & 32.3\\
\midrule
rep Yolo w/ SAM2 & 43.3  & 51.9 & 39.9 & 29.3 & 46.4 & 40.3 & 32.7\\
$\Rightarrow$ Streaming & 44.2 & 50.9 & 39.7 & 37.9 & 45.0 & 44.3 & 34.1 \\
Diversity only & 38.7 & 40.4 & 38.8 & 40.0 & 38.7 & 40.9 & 33.7 \\
Relevance only & 42.0 & 48.5 & 39.9 & 27.6 & 44.0 & 37.3 & 31.8 \\ 

\bottomrule
\end{tabu}
\vspace{-8pt}
\caption{\textbf{Ablation studies on LVBench.} \ApproachName~defaults to global mode with YOLO for efficiency, while SAM2 performs similarly.
Global mode excels in long-context tasks, and streaming mode in temporally-intensive tasks. 
Combining Relevance and Diversity in attention learning achieves the best performance.}
\vspace{-2ex}
\label{tab:lvbench-ablations}
\end{table}

\begin{figure}[t]
    \centering
    \includegraphics[scale = 0.47]{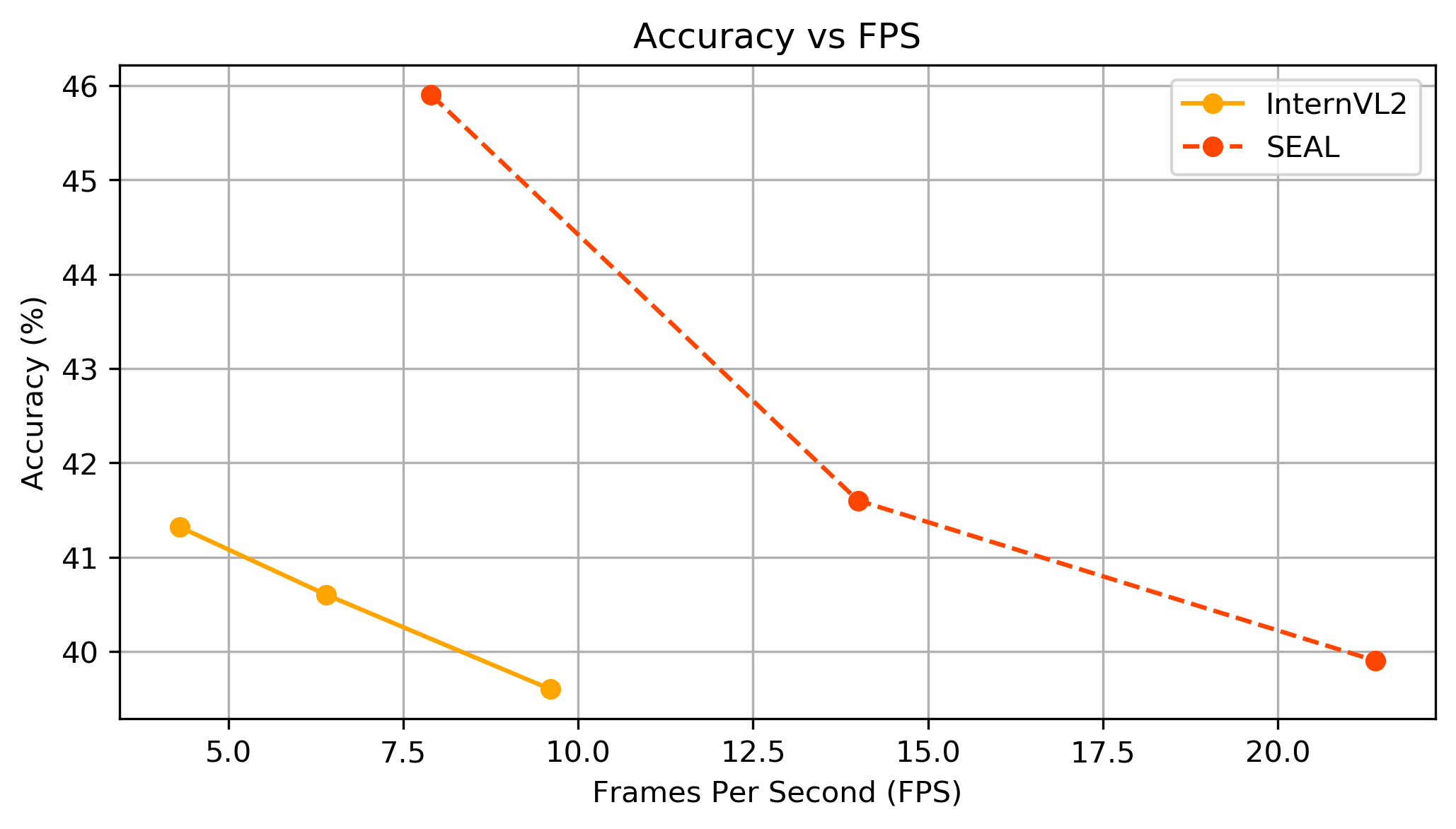}
    \vspace{-8pt}
    \caption{{\bf Accuracy vs. efficiency trade-off on LVBench}. SEAL runs 2-3x faster than InternVL2 at the same accuracy, and is more accurate when compared at the same FPS.}
    \label{fig:complexity}
    \vspace{-1ex}
\end{figure}

 \noindent {\bf Computational Complexity}. We provide an accuracy-efficiency trade-off comparison in Fig~\ref{fig:complexity} where both SEAL and InternVL2 are with a varying number of tokens. FPS is calculated during inference that includes all the processing time from the raw video frames.
SEAL reduces the reliance on densely sampled tokens and uses subject-level and motion aware representations/tokens to improve efficiency while maintain good accuracy. From Fig~\ref{fig:complexity}, at 10 FPS, SEAL is about 5\% more accurate than InternVL2. When comparing at the same accuracy of 41.5\% and 40\%, SEAL is about 3x and 2x faster than InternVL2, respectively.
\section{Conclusion and Future Work}

We propose \ApproachName, a novel unified representation for long video understanding that addresses computational complexity, temporal redundancy, and cross-task generalization. 
\ApproachName~leverages semantic decomposition to break videos into scene, object, and action tokens, reducing redundancy and enabling efficient processing. It incorporates attention learning to balance query relevance and token diversity, enhancing performance across diverse tasks. \ApproachName~achieves state-of-the-art results on benchmarks like MovieChat-1K, LVBench, and Ego4D-NLQ, demonstrating its versatility and effectiveness for long video understanding.

\noindent
{\bf Limitation.} The Attention Learning module is bounded by the memory constraint for the QP solver, making it not fully end-to-end trainable. For future work, we plan to integrate Attention Learning in our streaming mode for full end-to-end learning. Developing special prediction heads to solve causal reasoning \cite{li2022from}, where MLLM heads showed limitations, is also an interesting area for future exploration.

\noindent
{\textbf{Acknowledgements.} This work was partially supported by the Office of Naval Research (award \#N00014-23-1-2417). Any opinions, findings, and conclusions or recommendations expressed in this material are those of the authors and do not necessarily reflect the views of ONR.}

{
    \small
    \bibliographystyle{ieeenat_fullname}
    \bibliography{main}
}

\clearpage
\setcounter{page}{1}
\maketitlesupplementary
\appendix

\section{Additional Ablations}
\noindent {\bf Streaming Window Size}. Table~\ref{tab:lvbench-ablations-window} demonstrates the impact of different window sizes on performance in streaming mode. This ablation experiment simulates the behavior of streaming mode under varying memory constraints. The results show that the performance of streaming mode is optimal when the window size is set to 1000, demonstrating its ability to effectively balance memory usage and accuracy under this configuration.

\begin{table}[ht]
\footnotesize\centering
\begin{tabu} to 0.47\textwidth {@{}cX[2cm c]*{6}{X[c]}@{}}
\toprule
\textbf{Model} & \textbf{Overall} & \textbf{KIR} & \textbf{EU} & \textbf{Sum} & \textbf{ER} & \textbf{Rea} & \textbf{TG} \\
\midrule
{\bf Ours - 500} & 42.7   & 50.5 & 38.9 & 39.7 & 43.4 & 41.8 & 32.7 \\
{\bf Ours - 1000} & 44.2 & 50.9 & 39.7 & 37.9 & 45.0 & 44.3 & 34.1 \\
{\bf Ours - 2000} & 42.7   & 52.9 & 40.3 & 20.7 & 43.7 & 37.8 & 29.5 
\\
\bottomrule
\end{tabu}
\caption{\textbf{Ablation studies of Streaming Window Size on LVBench.} Streaming mode performs best when the window size is set to 1000. }
\label{tab:lvbench-ablations-window}
\end{table}

\noindent {\bf Partially-Observed Videos}. Table~\ref{tab:lvbench-ablations-partial}  presents the performance of different methods when only a portion of video is accessible including scenarios where only the first half or quarter of the video is available. This experiment simulates streaming mode, where the model receives only a portion of the video as input, evaluating its ability to answer questions under such constraints. The results show that our method significantly outperforms the uniform sampling approach of the InterVL2-40B model, highlighting the effectiveness of our relevant and diverse tokens.

\begin{table}[ht]
\footnotesize\centering
\begin{tabu} to 0.47\textwidth {@{}cX[2cm c]*{7}{X[c]}@{}}
\toprule
\scriptsize{\textbf{Model}} & \scriptsize{\textbf{Observed}} & \scriptsize{\textbf{Overall}} & \scriptsize{\textbf{KIR}} & \scriptsize{\textbf{EU}} & \scriptsize{\textbf{Sum}} & \scriptsize{\textbf{ER}} & \scriptsize{\textbf{Rea}} & \scriptsize{\textbf{TG}} \\
\midrule
InterVL2-40B & 1 & 39.6 & 43.4 & 39.7 & 41.4 & 37.4 & 42.5 & 31.4 \\
{\bf SEAL (Ours)} & 1 & 45.9   & 51.5 & 41.3 & 39.7 & 47.9 & 43.3 & 32.3\\
InterVL2-40B & 1/2 & 35.7 & 34.4 & 34.2 & 37.9 & 35.7 & 40.3 & 28.2 \\
{\bf SEAL (Ours)} & 1/2 & 41.6   & 50.9 & 37.9 & 41.4 & 41.9 & 39.8 & 29.5 \\
InterVL2-40B & 1/4
& 35.6 & 36.4 & 33.8 & 34.4 & 34.1 & 37.5 & 27.3 \\
{\bf SEAL (Ours)} & 1/4 & 39.3   & 40.9 & 38.6 & 31.0 & 41.1 & 34.8 & 33.2 \\
\bottomrule
\end{tabu}
\caption{\textbf{Ablation studies of prediction with partially-observed videos on LVBench.} When only partial videos are visible, the performance of traditional uniform sampling drops significantly, while our method shows more reasonable results.}
\label{tab:lvbench-ablations-partial}
\end{table}

\begin{figure*}
    \centering
    \includegraphics[width = 0.9 \textwidth]{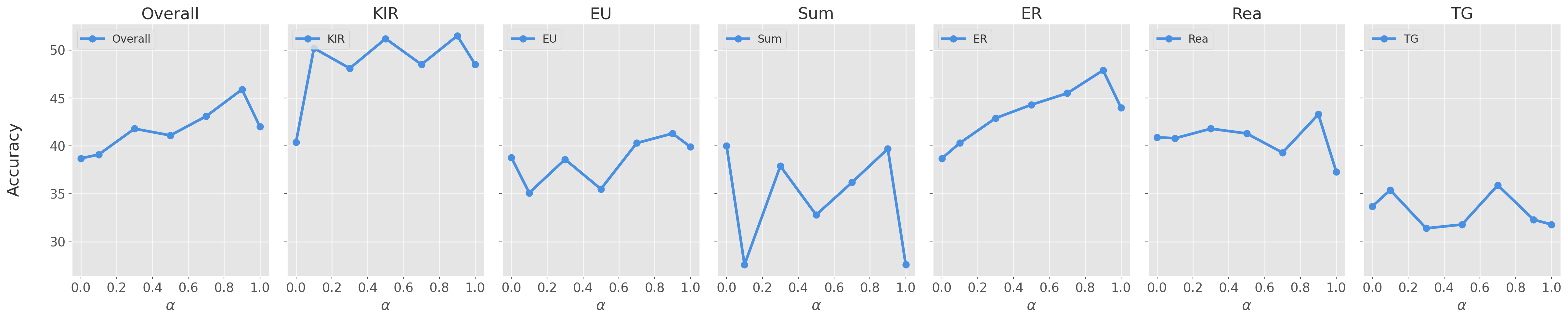}
    \caption{\textbf{Ablation studies of different values of $\alpha$ on LVBench.} $\alpha = 0.9$ achieves the best performance across different tasks except for temporal grounding (TG).
}
    \label{fig:alpha}
\end{figure*}
\begin{figure*}[ht]
    \centering
    \includegraphics[width = 0.95\textwidth]{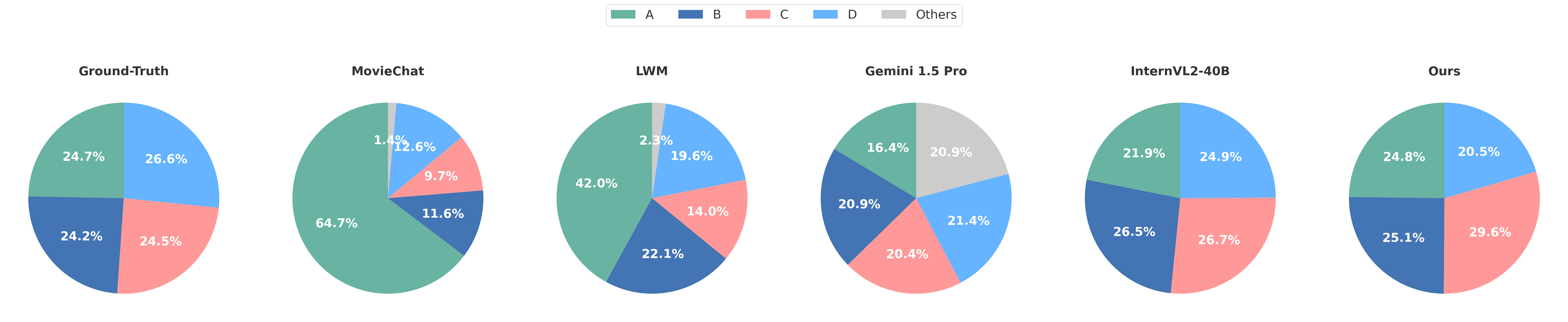}
    \caption{\textbf{Distribution of answers generated by different models.} The answers from InterVL2-40B and our method are the closest to the ground truth distribution. }
    \label{fig:supp_fig1}
\end{figure*}

\begin{table*}[ht]
\centering

\begin{tabular}{l|ccccccc}
\toprule
Model & Sports & Documentary & Event Record & Lifestyle & TV Show & Cartoon & Overall \\  \midrule

\textcolor{gray}{Random predictions} & \textcolor{gray}{27.5}   & \textcolor{gray}{25.4}        & \textcolor{gray}{23.3}         & \textcolor{gray}{23.3}      & \textcolor{gray}{25.6}    & \textcolor{gray}{25.8}    & \textcolor{gray}{25.1}    \\ 

\textcolor{gray}{Random tokens} & \textcolor{gray}{25.4}   & \textcolor{gray}{25.9}        & \textcolor{gray}{25.6}         & \textcolor{gray}{26.2}      & \textcolor{gray}{24.4}    & \textcolor{gray}{21.6}    & \textcolor{gray}{24.8}    \\ 

\textcolor{gray}{Human} & \textcolor{gray}{96.3}   & \textcolor{gray}{89.8}        & \textcolor{gray}{87.4}         & \textcolor{gray}{98.4}      & \textcolor{gray}{97.2}    & \textcolor{gray}{95.8}    & \textcolor{gray}{94.4}    \\ 
\hdashline
InternVL2-40B & 43.5   & 45.2        & 38.9         & 41.6      & 32.8    & 36.4    & 39.5    \\ 
Qwen2-VL-72B  & 43.0   & 42.6        & 40.8         & 41.0      & 42.0    & 38.9    & 41.3    \\  
\textbf{SEAL (Ours)} & 49.2 & 49.2 & 48.1 & 46.7 & 44.4& 39.2 & 45.9
\\
\bottomrule
\end{tabular}
\caption{\textbf{Evaluation across different video categories on LVBench.} Comparing our method with baselines and state-of-the-art approaches on different video categories. Our method consistently outperforms state-of-the-art approaches on all categories. Although our method has made substantial improvements over lower-bound baselines (Random tokens and Random predictions), it still has a significant gap compared with the upper-bound baseline of human performance.}
\label{tab:lvbench_results}
\end{table*}
\begin{figure*}[ht]
    \centering
    \includegraphics[width = 0.8\textwidth]{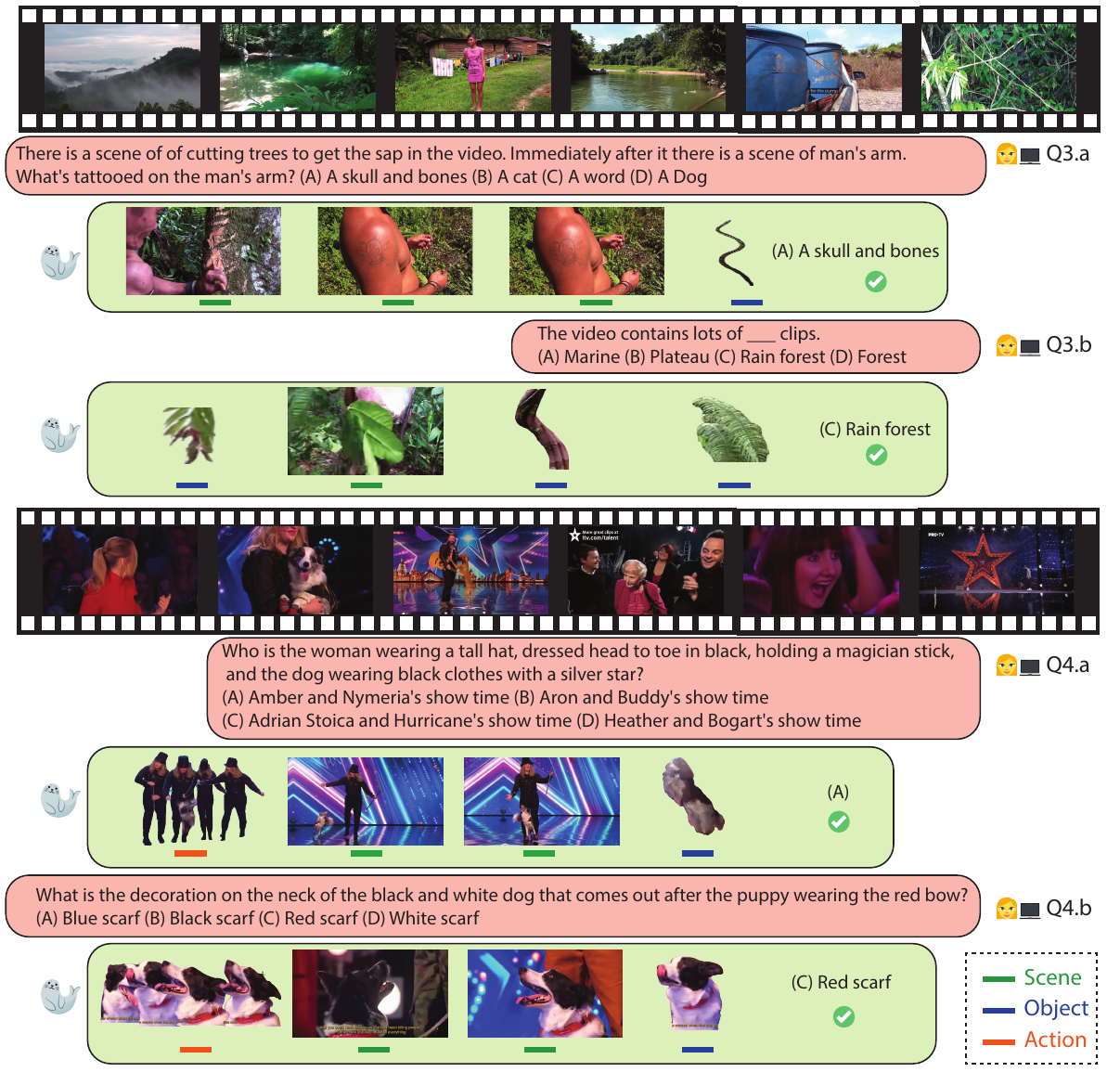}
    \caption{{\bf Additional qualitative results on LVBench}.\ApproachName~ attends to relevant entities such as ``tattoo'' and ``man's arm'' (Q3.a), different ``rain forest plants'' and ``rain forest leaves'' (Q3.b), ``tall hat woman'',  ``dog'', and ``performing" activity (Q4.a), `` black and white dog" and its activity (Q4.b) and correctly answers these questions. However, the answers provided by InterVL2-40B are C, D, C, C for Q3.a, Q3.b, Q4.a, and Q4.b, respectively. This indicates that InterVL2-40B fails to capture key information such as "tattoo", "rainforest", and important details about the main characters in the performance. }
    \label{fig:fig_sup}
\end{figure*}
\noindent {\bf Ablation on Different $\alpha$.}
Figure~\ref{fig:alpha} shows the performance trends across various categories for different values of $\alpha$. $\alpha = 0.9$ achieves the best overall trade-off, reaching peak with the highest overall accuracy of \textbf{45.9}. Conversely, extreme values like $\alpha = 0.0$ or $1.0$ lead to declines in several metrics, highlighting that both diversity and relevance are essential. Therefore, $\alpha = 0.9$ is the optimal choice for experiments, delivering peak performance and a well-rounded balance across all categories.

\noindent {\bf Effectiveness of Encoder for Relevance}. Table \ref{tab:blip} presents the relevance results computed using the BLIP (Base), CLIP (ViT-L/14) and the BLIP2 (Large)  models. The results demonstrate that stronger models achieve higher effectiveness in computing relevance scores, leading to significant performance gains for SEAL.

\begin{table}[h]
\footnotesize\centering
\begin{tabu} to 0.47\textwidth {@{}lX[2cm c]*{6}{X[c]}@{}}
\toprule
\textbf{Model} & \textbf{Overall} &
\textbf{KIR} & \textbf{EU} & \textbf{Sum} & \textbf{ER} & \textbf{Rea} & \textbf{TG} \\
\midrule
{\bf SEAL w/ BLIP2} & 45.9   & 51.5 & 41.3 & 39.7 & 47.9 & 43.3 & 32.3\\
{\bf SEAL w/ CLIP}  & 42.9 & 48.5 & 38.6 & 36.2 & 46.2 & 36.3 & 32.7 \\
{\bf SEAL w/ BLIP} & 40.5 & 41.9 & 38.6 & 39.7 & 40.5 & 47.2 & 32.7\\

\bottomrule
\end{tabu}
\caption{Comparison of BLIP2 with other methods on LVBench.}
\label{tab:blip}
\end{table}

\section{Additional Results and Discussions}

\noindent {\bf Comparison with LVU methods on LVBench}. We provide additional comparison with LVU methods on LVBench in Table~\ref{tab:representation}. For a fair comparison, we follow those methods to use a 7B LLM. SEAL maintains superior performance with a much smaller LLM (7B), demonstrating the effectiveness of our proposed method.

\begin{table}[h]
\footnotesize\centering
\begin{tabu} to 0.47\textwidth {@{}lX[2cm c]*{6}{X[c]}@{}}
\toprule
\textbf{Model} & \textbf{Overall} &
\textbf{KIR} & \textbf{EU} & \textbf{Sum} & \textbf{ER} & \textbf{Rea} & \textbf{TG} \\
\midrule

\textbf{MovieChat~\cite{song2023moviechat}} & 22.5 & 25.9 & 23.1 & 17.2 & 21.3 & 24.0 & 22.3\\
\textbf{TimeChat~\cite{ren2024timechat}} & 22.3 & 25.9 & 21.7 & 24.1 & 21.9 & 25.0 & 22.7\\
\textbf{MA-LLM~\cite{he2024ma}} & 24.5 & 25.4 & 25.8 & 22.4 & 22.3 & 26.9 & 21.8 \\
{\bf SEAL (7B)} & 36.6 & 44.3 & 33.7 & 27.6 & 36.9 & 32.8 & 30.9\\

\bottomrule
\end{tabu}
\caption{Comparison with other long video representations on LVBench.}
\label{tab:representation}
\end{table}

\noindent {\bf Number of different tokens}. The subset of tokens is learned as an optimization problem in Section 3.2, and the composition of tokens varies on different inputs and tasks.
The averaged percentages of scene, object, and action tokens are 62.5\%, 26.1\%, 11.4\% on LVBench,  54.3\%, 25.6\%, 20.1\% on Moviechat,  88.5\% scene tokens and 11.5\% action tokens on Ego4d-NLQ. Since Ego4D-NLQ is a temporal localization task, we only utilize scene and action tokens.

\noindent {\bf Result Analysis}. We evaluated the distribution of answers generated by different models, following~\cite{wang2024lvbench}, as shown in Figure~\ref{fig:supp_fig1}. The Ground-Truth exhibits a fairly balanced distribution among  A, B, C, and D, indicating a well-distributed dataset where no single category is disproportionately represented. MovieChat and LWM shows a dominance of category A, with significantly smaller contributions from other categories, suggesting a lack of diversity in predictions. In Gemini 1.5 Pro, the ``Others'' category is significantly high, indicating that Gemini 1.5 Pro produces a notable number of unrelated outputs. Our method demonstrates a distribution close to Ground-Truth, showing strong generalization and robustness.

We evaluate performance across various video categories in Table~\ref{tab:lvbench_results}. The Human benchmark achieves the highest accuracy across all categories, with an overall accuracy of 94.4\%. Ours method achieves an overall accuracy of 45.9\%, representing a clear improvement over InternVL2-40B  and Qwen2-VL-72B. This demonstrates the our model's ability to generalize better across different video categories. However, the performance in the Cartoon category shows less improvement relative to other categories, indicating potential challenges in handling stylized or abstract visual content. While our method shows clear improvements over existing models, there remains a substantial gap with the Human benchmark across all categories. Further study is needed to enhance the model's understanding of long videos.

\section{Additional Qualitative Results}

Figure~\ref{fig:fig_sup} presents additional qualitative results of \ApproachName~on LVBench, showcasing its ability to focus on relevant semantic tokens and provide correct answers to various types of questions. Compared to InterVL2-40B, \ApproachName~effectively attends to critical entities, such as ``tattoo'' and ``man's arm'' (Q3.a), distinct ``rain forest plants'' and ``rain forest leaves'' (Q3.b), ``tall hat woman'', ``dog'', and the ``performing'' activity (Q4.a), as well as the ``black and white dog'' and its activity (Q4.b), resulting in accurate answers. In contrast, the answers provided by InterVL2-40B are C, D, C, and C for Q3.a, Q3.b, Q4.a, and Q4.b, respectively. This highlights that InterVL2-40B struggles to capture key information, such as ``tattoo'', ``tall hat woman'', and to distinguish ``rain forest'' from ``forest'' (InterVL2-40B chose ``forest'' failing to capture subtle features related to ``rain forest''), as well as critical details about the main characters and activities in the scene. These results underscore the superior reasoning capabilities of \ApproachName.

\section{Additional Implementation Details}

\subsection{Token Extraction}

\noindent \textbf{Scene token}. We use the full frames to represent scene tokens. The full frames or clips are fed into encoders (2D or 3D CNN/ViT) to extract the token embeddings. For MovieChat and LVBench, we use a frame-based 2D encoder~\cite{fang2023eva} and~\cite{internvl}. For the Ego4D-NLP dataset, we follow~\cite{mu2024snag} and use a 3D clip-based encoder~\cite{kevin2022egovlp} for processing 23-frame clips.

\noindent \textbf{Object token}. For object tokens, we extract masks using from SAM2~\cite{SAM2} {\em Automatic Mask Generator}. For mask prediction, we sample \(64 \times 64\) points per image for dense and uniform coverage, with a batch size of 128 points to balance computational efficiency and memory usage. Predicted masks are filtered using a quality threshold of \texttt{pred\_iou\_thresh=0.88}, retaining only masks with high predicted IoU scores, and a stability score threshold of \texttt{stability\_score\_thresh=0.92}, ensuring the robustness of masks under varying binarization cutoffs. To calculate the stability score, the cutoff is shifted by \texttt{stability\_score\_offset=0.99}. Non-maximal suppression (NMS) is applied with an IoU threshold of \texttt{box\_nms\_thresh=0.7} to remove redundant masks. We do not employ additional cropping layers (\texttt{crop\_n\_layers=0}). We extract features based on the mask's bounding box, expand it by 2x to include additional contextual information, and use the same encoder as the scene token for different datasets. We set \(N_{\text{key}}=128\) for MovieChat and \(N_{\text{key}}=64\) for Ego4D-NLP and LVBench datasets.

\noindent \textbf{Action token}.
For the Ego4D-NLP and LVBench datasets, we use YOLOv10-X~\cite{Wang2024YOLOv10RE} with BoT-SORT~\cite{Aharon2022BoTSORTRA} for extracting action tracklets. For MovieChat, we employ NetTrack~\cite{nettrack} for action tracklets. We set \(L_{\text{min}} = 8\) and \(L_{\text{max}} = 16\) for MovieChat, while for Ego4D-NLP, we set \(L_{\text{min}} = 16\) and \(L_{\text{max}} = 32\). For LVBench, since the action token encoder~\cite{internvl} is a frame-based encoder, we use the middle frame of all action tracklets as the action token candidates.

In Attention Learning stage, we sample in total 256 tokens for MovieChat, 200 / 450 tokens for Ego4D-NLP and 16 tokens for LVBench.  Note that since the task of Ego4D-NLP is temporal grounding, we only used action tokens and scene tokens to ensure temporal consistency.

\subsection{LLM Heads and LLM-based Evaluation}
For the MovieChat dataset, we provide the large language model with the following prompt for the Video QA task: 
\begin{quote}
\texttt{"You are able to understand the visual content that the user provides. Follow the instructions carefully and explain your answers."}
\end{quote}
For the LVBench dataset, given a question and options, we use the prompt for the Video QA multiple choice task:
\begin{quote}
\texttt{"Please select the best answer from the options above and directly provide the letter representing your choice without giving any explanation."}
\end{quote}
Following~\cite{song2023moviechat}, we use LLM-Assisted Evaluation for the video question-answering task when evaluating MovieChat dataset. Given the question, the correct answer, and the predicted answer provided by different methods, the LLM assistants should return a True or False judgment along with a relative score ranging from 0 to 5. we provide the large language model with the following prompt:
\begin{quote}
\texttt{"Provide your evaluation only as a yes/no and score where the score is an integer value between 0 and 5, with 5 indicating the highest meaningful match."}
\end{quote}    

\end{document}